\pdfoutput=1

\documentclass[11pt]{article}
\usepackage{svg}
\usepackage{multirow}
\usepackage{hyperref}
\usepackage{svg-extract}
\usepackage{breakurl}
\usepackage[splitrule]{footmisc}
\usepackage{algorithmic}

\usepackage[]{acl}
\usepackage{listings}
\usepackage{times}
\usepackage{latexsym}

\usepackage[T1]{fontenc}

\usepackage[utf8]{inputenc}

\usepackage{microtype}

\usepackage{inconsolata}

\usepackage{graphicx}

\usepackage{paralist} 

\newcommand{\frameworkname}{{MapQaTor}~}
\newcommand{\frameworknamenospace}{{MapQaTor}}

\title{MapQaTor: An Extensible Framework for\\ Efficient Annotation of Map-Based QA Datasets}

\author{
Mahir Labib Dihan$^1$, \ Mohammed Eunus Ali$^{1,2}$, \ Md Rizwan Parvez$^3$ \\
$^1$Department of Computer Science and Engineering \\
Bangladesh University of Engineering and Technology (BUET) \\
$^2$Faculty of Information Technology, \\ Monash University, Melbourne, Australia \\
$^3$Qatar Computing Research Institute (QCRI) \\
\{mahirlabibdihan, mohammed.eunus.ali\}@gmail.com, mparvez@hbku.edu.qa \\
\raisebox{-1.2pt}{\includegraphics[scale=0.05]{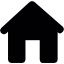}}\,\hspace{2pt}\url{https://mapqator.github.io/project/}
}

\makeatletter
\let\@oldmaketitle\@maketitle%
\renewcommand{\@maketitle}{\@oldmaketitle%
  \vspace{10pt}
  \centering
  \includegraphics[width=0.98\linewidth]{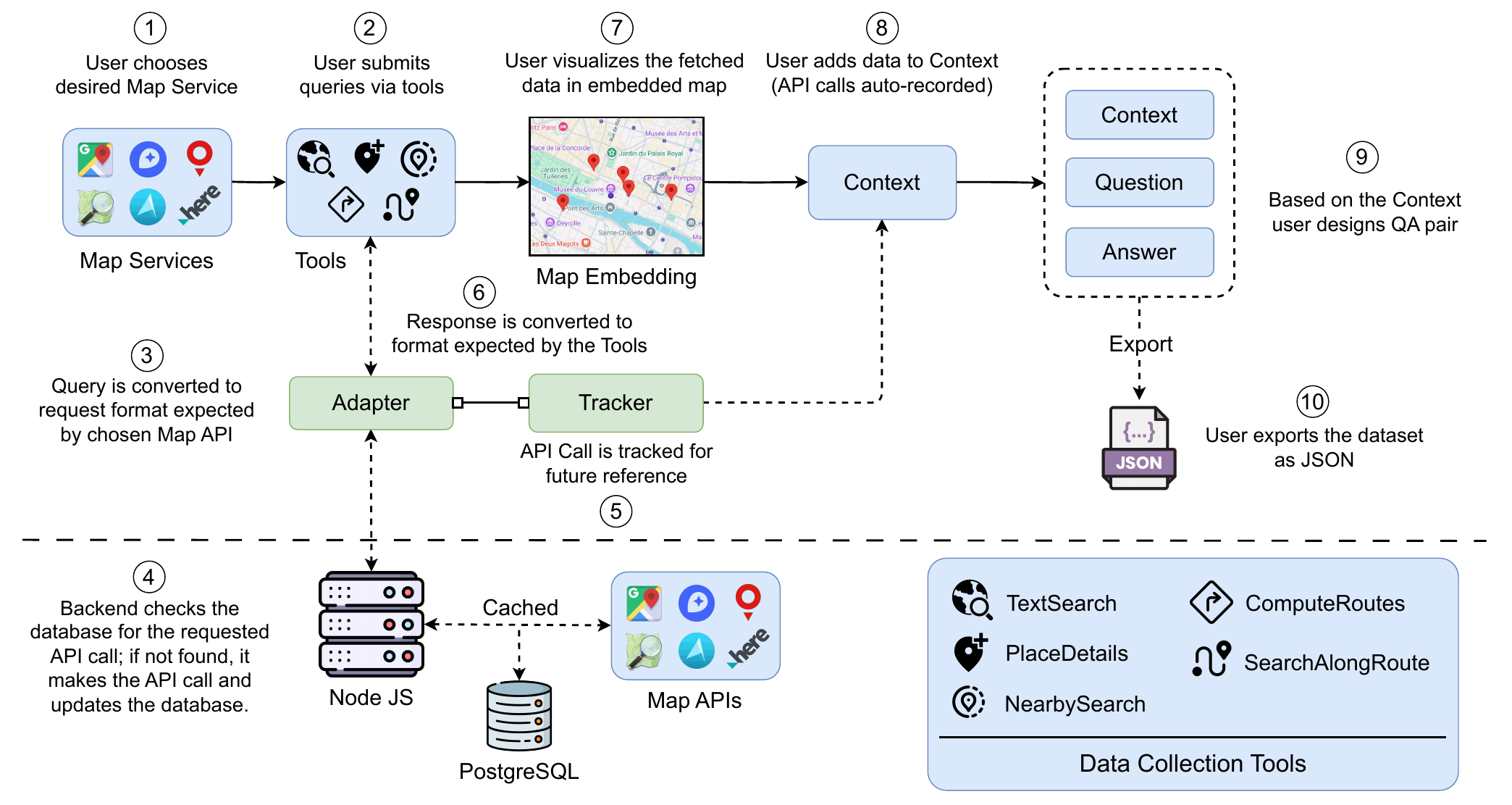}
  \vspace{-5pt}
  \captionof{figure}{
    Overview of the annotation and visualization process of \frameworkname.
  }
  \label{fig:mapqator-overview}
  \vspace{15pt}
 }
\makeatother

\begin{document}
\maketitle

\begin{abstract}

Mapping and navigation services like Google Maps, Apple Maps, OpenStreetMap, are essential for accessing various location-based data, yet they often struggle  to handle natural language geospatial queries. Recent advancements in Large Language Models (LLMs) show promise in question answering (QA), but creating reliable geospatial QA datasets from map services remains challenging. We introduce MapQaTor, an extensible open-source framework that streamlines the creation of reproducible, traceable map-based QA datasets. MapQaTor enables seamless integration with any maps API, allowing users to gather and visualize data from diverse sources with minimal setup. By caching API responses, the platform ensures consistent ground truth, enhancing the reliability of the data even as real-world information evolves. MapQaTor centralizes data retrieval, annotation, and visualization within a single platform, offering a unique opportunity to evaluate the current state of LLM-based geospatial reasoning while advancing their capabilities for improved geospatial understanding. Evaluation metrics show that, MapQaTor speeds up the annotation process by at least 30 times compared to manual methods, underscoring its potential for developing geospatial resources, such as complex map reasoning datasets. The website is live at: \url{https://mapqator.github.io/} and a demo video is available at: \url{https://youtu.be/bVv7-NYRsTw}.

\end{abstract}.


\begin{table*}[h]
\centering
\begin{tabular}{|c|p{3.2cm}|p{6cm}|}
\hline
\textbf{Tool} & \textbf{API Provider} & \textbf{API Endpoint} \\
\hline
\multirow{7}{*}{Text Search} & \multirow{2}{*}{Google Maps} & \href{https://developers.google.com/maps/documentation/places/web-service/text-search}{Text Search (New) | Places API} \\ 
\cline{3-3} & & \href{https://developers.google.com/maps/documentation/places/web-service/search-text}{Text Search | Places API} \\
\cline{2-3} & OpenStreetMap & \href{https://nominatim.org/release-docs/develop/api/Search/}{Search queries | Nominatim} \\ 
\cline{2-3} & Mapbox  & \href{https://docs.mapbox.com/api/search/search-box/}{Suggest | Search Box API} \\ 
\cline{2-3} & TomTom  & \href{https://developer.tomtom.com/search-api/documentation/search-service/points-of-interest-search}{Point of Interest Search} \\ 
\cline{2-3} & HERE  & \href{https://www.here.com/docs/bundle/geocoding-and-search-api-developer-guide/page/topics/endpoint-discover-brief.html}{Discover | Geocoding and Search}  \\ 
\cline{2-3} & Azure Maps  & \href{https://learn.microsoft.com/en-us/rest/api/maps/search/get-search-fuzzy?view=rest-maps-1.0&tabs=HTTP}{Search - Get Search Fuzzy}  \\ 
\hline
\multirow{6}{*}{Place Details} & Google Maps & \href{https://developers.google.com/maps/documentation/places/web-service/place-details}{Place Details (New) | Places API} \\ 
\cline{2-3} & OpenStreetMap & \href{https://nominatim.org/release-docs/develop/api/Details/}{Place details | Nominatim} \\ 
\cline{2-3} & Mapbox & \href{https://docs.mapbox.com/api/search/search-box/}{Retrieve | Search Box API} \\ 
\cline{2-3} & TomTom & \href{https://developer.tomtom.com/search-api/documentation/place-by-id-service/place-by-id}{Place by ID} \\ 
\cline{2-3} & HERE & \href{https://www.here.com/docs/bundle/geocoding-and-search-api-developer-guide/page/topics/endpoint-lookup-brief.html}{Lookup | Geocoding and Search} \\ 
\cline{2-3} & Azure Maps  & \href{https://learn.microsoft.com/en-us/rest/api/maps/search/get-search-fuzzy?view=rest-maps-1.0&tabs=HTTP}{Search - Get Search Fuzzy}  \\ 
\hline
\multirow{2}{*}{Nearby Search} & Google Maps & \href{https://developers.google.com/maps/documentation/places/web-service/nearby-search}{Nearby Search (New) | Places API} \\ 
\cline{2-3} & TomTom & \href{https://developer.tomtom.com/search-api/documentation/search-service/nearby-search}{Nearby Search} \\
\hline
\multirow{3}{*}{Compute Routes} & Google Maps & \href{https://developers.google.com/maps/documentation/routes/compute_route_directions}{Get a route | Routes API} \\ 
\cline{2-3} & OpenStreetMap & \href{https://docs.graphhopper.com/#tag/Routing-API}{Routing API | GraphHopper} \\ 
\cline{2-3} & TomTom & \href{https://developer.tomtom.com/routing-api/documentation/tomtom-maps/calculate-route}{Calculate Route} \\
\hline
\multirow{2}{*}{Search Along Route} & Google Maps & \href{https://developers.google.com/maps/documentation/places/web-service/search-along-route}{Search along route} \\ 
\cline{2-3} & TomTom & \href{https://developer.tomtom.com/search-api/documentation/search-service/along-route-search}{Along Search Route} \\ 
\hline
\end{tabular}
\caption{Current API Support for Data Collection Tools in \frameworkname}
\label{tab:api_support}
\end{table*}

\section{Introduction}
In recent years, mapping and navigation services have transformed the way individuals access and interact with location-based information. Platforms such as \href{https://mapsplatform.google.com/}{Google Maps} and \href{https://www.apple.com/au/maps/}{Apple Maps} have become essential tools, providing users with features like route planning, nearby points of interest (POIs), and contextual data, including reviews and operating hours. However, while these services offer extensive geospatial data, they often struggle to understand and process natural language queries. This limitation hampers their effectiveness for users seeking to obtain specific information or engage in more complex question-answering (QA) tasks.

Recent advancements in multi-agent and tool-augmented large language models (LLMs) demonstrate significant promise for complex reasoning, decision-making, and generation tasks across various application domains, including those that interact with domain-specific tools such as maps \cite{liu2023agentbench, qin2023toolllm}. Notable tasks like WebArena \cite{zhou2023webarena} and VisualWebArena \cite{koh2024visualwebarena} have been proposed with practical real-life applications involving map usage. However, despite these developments, there remains no straightforward method for LLMs to access the vast databases of map services. Currently, there are no dedicated platforms designed to efficiently annotate language-map reasoning tasks, such as question answering. This gap leads to significant challenges in creating reliable datasets for training and evaluating LLMs for geospatial reasoning tasks, as many existing approaches rely on manual data collection methods that result in inconsistencies, lack of reproducibility, and difficulties in tracking the origins of information.

To address these issues, we present \frameworknamenospace, a web application designed to streamline the creation of map-based QA datasets. MapQaTor empowers researchers to seamlessly integrate with any map API, enabling them to gather, visualize, and annotate geospatial data from desired map API with minimal setup. By caching API responses, the platform ensures a consistent ground truth, which enhances the reliability of the datasets, even as real-world information evolves over time.

In summary, in this demo we have made the following key contributions:
\begin{compactenum}
\item We propose a novel framework, \frameworknamenospace, first of its kind, which simplifies the creation of reproducible map-based QA datasets and reduces reliance on manual data collection through its extensible architecture, enabling seamless integration with any map API (e.g., Google Maps, Apple Maps, OpenStreetMap).
\item We provide visualization tools that facilitate better understanding and annotation of geospatial information.
\item  We implement caching of API responses to ensure a consistent ground truth, enhancing the reliability of QA tasks over time.
\item We evaluate \frameworkname to estimate its usefulness and efficiency. 
\end{compactenum}

We have published the code on GitHub\footnote{\url{https://github.com/mapqator/}} under the Apache 2 license.

\begin{figure*}[t]
    \centering
     \includegraphics[width=1\linewidth]{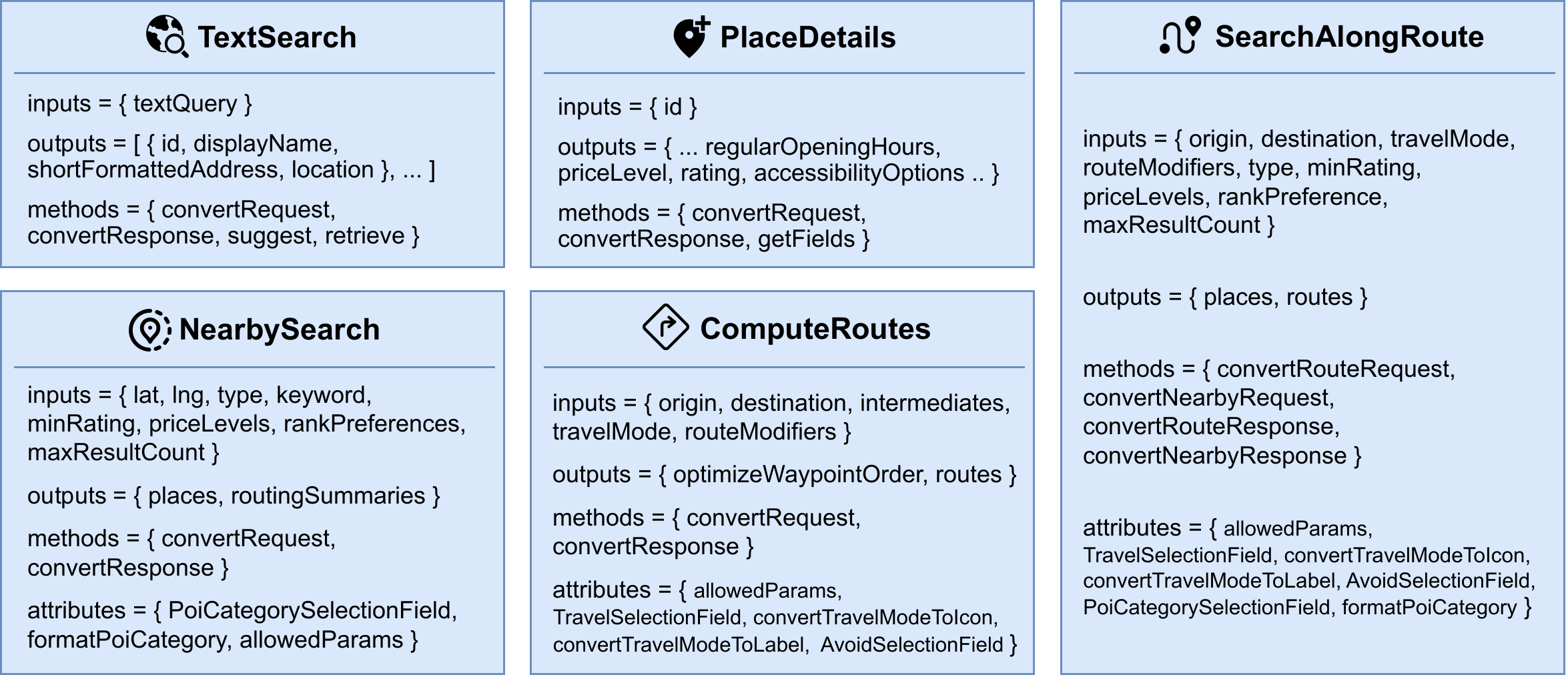}
    \caption{Standardized schema for data collection tools, unifying inputs, outputs, methods, and attributes.}
        \label{fig:architecture}
\end{figure*}

\section{\frameworkname}\label{sec:system}

\frameworkname\ is a web-based platform designed to streamline the creation of reproducible, map-based question-answering (QA) datasets that can be used to evaluate and advance the geospatial reasoning abilities of large language models (LLMs). By integrating with any map API, \frameworkname enables users to efficiently gather, annotate, and visualize map data to support complex, location-based QA tasks. This section details the main components of the platform, its architecture, and its unique features. Figure~\ref{fig:mapqator-overview} outlines the proposed framework, which enables users to interact with map APIs by submitting queries, processing responses, and visualizing data. The framework allows users to design question-answer pairs and export the dataset in JSON format for downstream applications. The whole working flow is shown using ten key steps. 

\subsection{Context Designer}\label{sec:data-collection-tools}
The core function of MapQaTor is to generate Context\footnote{Context refers to the data and information necessary to design a QA pair, ensuring that the answer to each question exists within the context.} using data collection tools, enabling structured and efficient QA pair creation. 

\subsubsection{Data Collection Tools}
\frameworkname's data collection framework (Figure \ref{fig:architecture}) integrates five modular tools—Text Search, Place Details, Nearby Search, Compute Routes, and Search Along Route—to unify diverse map API functionalities under a standardized interface. Each tool follows a consistent design pattern:

\begin{compactitem}
\item {Inputs}: User-defined parameters (e.g., location coordinates, filters, natural language queries).
\item {Outputs:} Structured API responses (e.g., places, routes, metadata) normalized for downstream tasks.
\item Context Integration: All inputs, raw API outputs, and processed data are stored as reusable Context, preserving traceability, and enabling QA generation.
\end{compactitem}

\noindent The tools abstract API-specific complexities through configurable adapters while maintaining provider flexibility. Below, we outline their roles and workflows, with visual examples.

\noindent\textbf{Text Search:}
Allows users to search for places by entering free-text queries (e.g., “Eiffel Tower” or “Starbucks near Central Park”). This tool leverages map API search capabilities to retrieve place names, addresses, and coordinates, making it efficient for locating points of interest (Figure \ref{fig:text-search}).

\noindent\textbf{Place Details:}
Fetches granular metadata (e.g., opening hours, accessibility) for a selected location (Figure \ref{fig:place-details}). It resolves API schemas into unified fields, supporting factual queries like “Does the Louvre Museum offer wheelchair access?”

\noindent\textbf{Nearby Search:} 
Finds points of interest (POIs) near a location (Figure~\ref{fig:nearby-places-form}). Users can filter by price tiers, ratings, and ranking logic, enabling spatial QA pairs like “List nearby restaurants of Eiffel Tower with at least a 4 rating.”

\noindent\textbf{Compute Routes:}
Generates navigation paths between locations (Figure \ref{fig:directions}), supporting multi-stop optimization and travel mode selection (e.g., driving, walking), with step-by-step instructions and route metrics.

\noindent\textbf{Search Along Route:}
Identifies POIs along a route (Figure \ref{fig:search-along-route-form}). Users specify filters and route parameters, enriching trip-planning contexts like “Find gas stations along Highway 1 from San Francisco to Los Angeles.”

\subsubsection{Context Management}
Each tool’s execution appends a Context entry containing:
\begin{compactitem}
\item Raw API Data: Original JSON responses for debugging and reproducibility.
\item Normalized Fields: Extracted attributes (e.g., coordinates, ratings) in a unified schema.
\item Metadata: Timestamps, API provider, and query parameters.
\end{compactitem}
This layered organization ensures flexibility: raw data supports provider-specific analysis, while normalized fields streamline QA generation.

\subsubsection{Impact on Reproducibility}
The architecture guarantees that identical queries produce the same structured outputs, even if the underlying API changes. For example, a Nearby Search for “restaurants near Louvre Museum” returns normalized fields like \texttt{rating}, \texttt{price}, and \texttt{coordinates}, regardless of whether Google Maps or OpenStreetMap is used. This consistency is critical for long-term dataset validity.



\subsubsection{Visualization Tools} \label{sec:visualization}


For visualizing geospatial data, \frameworkname utilizes the Google Maps JavaScript API\footnote{\url{https://developers.google.com/maps/documentation/javascript/overview}} to display places and routes directly on an embedded map. Users can view places as markers and visualize route paths (Figures \ref{fig:text-search}–\ref{fig:search-along-route-form}). To render routes, \frameworkname decodes polyline-encoded data from map APIs into latitude-longitude coordinates using polyline decoding algorithm \footnote{\url{https://developers.google.com/maps/documentation/routes/polylinedecoder}}, ensuring accurate visualization of complex routes. These visualization tools help users understand spatial relationships, facilitating the creation of precise and context-aware map-based questions.

\begin{figure}[t]
    \centering
    \fbox{\includegraphics[width=1\linewidth]{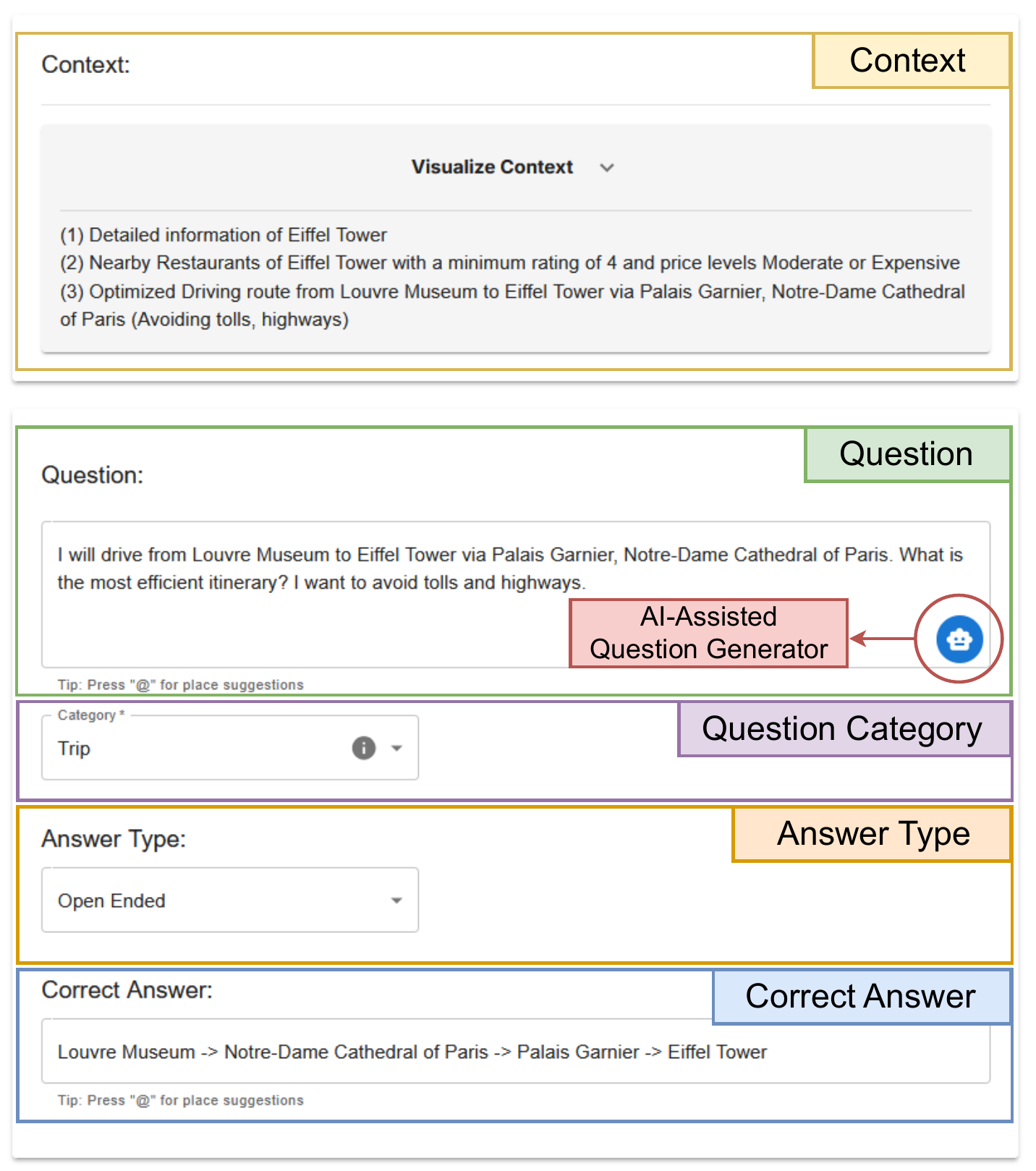}}
    \caption{QA design and annotation interface.}
    \label{fig:question-creation}
\end{figure}

\subsection{Question Design and Annotation}
The Question Design and Annotation feature in \frameworkname\ facilitates the creation and management of questions, enhancing the process of generating high-quality QA pairs (Figure \ref{fig:question-creation}). It supports four answer formats: Yes/No, Single Choice, Multiple Choice, and Open Ended, allowing users to select the format that best suits their needs.
Users can assign categories to each question, enabling better organization and retrieval based on thematic relevance. Also, while writing question/answer user will get Place Name suggestions to ensure consistency and uniqueness (Appendix \ref{sec:place-name-suggestion}). The system also supports AI-assisted question generation, leveraging Gemini-2.0-Flash \cite{Gemini2025} with few-shot prompting to automatically generate sample question from context, further enhancing the annotation process.
Once QA pairs are created, they can be evaluated using the Prompt Design Interface (see Appendix \ref{sec:prompt-designer}). This interface allows users to structure prompts, compare model's responses against ground truth, and assess the performance.

\subsection{Context Optimization}\label{sec:context-optimization}
\begin{figure}[t]
    \centering
     \includegraphics[width=1\linewidth]{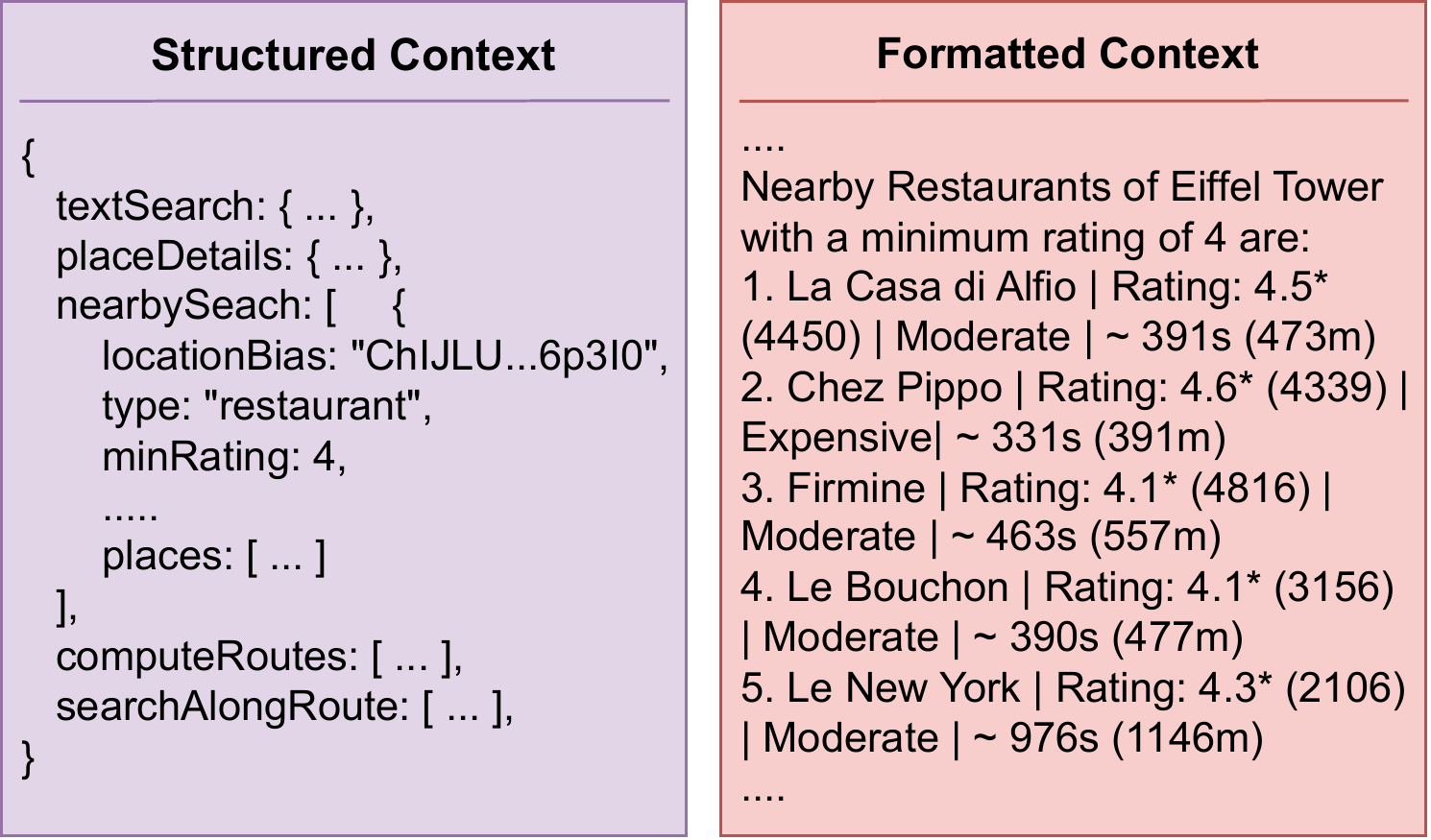}
    \caption{Comparison of structured and formatted context for improved readability and reduced size.}
        \label{fig:structured-vs-formatted-context}
\end{figure}

The structured context generated by MapQaTor’s data collection tools is often large and complex, containing detailed raw data and numerous metadata elements. While this structure is necessary to ensure complete traceability and data accuracy, it can be cumbersome when used directly in downstream tasks. To address this challenge, we convert the structured context into a more formatted context, which is a more compact, human-readable version (See figure \ref{fig:structured-vs-formatted-context}). This transformation retains the key information needed for evaluating LLMs for QA tasks, while eliminating unnecessary complexity. By simplifying the context, we significantly reduce token usage and improve processing efficiency, making it more suitable for large-scale evaluations and effective LLM-based analysis.

\subsection{API Extensibility}
New APIs can be integrated into \frameworkname\ by extending base tool classes (e.g., \texttt{NearbySearch}) and implementing abstract methods (e.g., \texttt{convertRequest}, \texttt{convertResponse}) as shown in Figure~\ref{fig:tomtom-adapter-layer}. Attributes like \texttt{PolCategorySelectionField} and \texttt{allowedParams} (Figure \ref{fig:architecture}) handle provider-specific UI elements, such as point-of-interest (POI) categories, which vary across APIs (e.g., Google Maps vs. OpenStreetMap). To date, \frameworkname\ has integrated 20 APIs from 6 providers (Table \ref{tab:api_support}), including both paid and free options. This modular design ensures adaptability to diverse map APIs while maintaining a consistent user experience. 

\subsection{Secure API Handling}  
\frameworkname’s backend securely mediates interactions between frontend tools (e.g., Nearby Search, Text Search) and third-party map APIs through two critical steps:  \\
\textbf{Tool-to-Backend Requests}:  
As shown in Figure~\ref{fig:tomtom-adapter-layer}, frontend tools send API-agnostic requests containing credential placeholders (e.g., \texttt{key:TOMTOM\_API\_KEY}) and provider-specific parameters.  \\
\textbf{API Key Injection}:  
The backend replaces placeholders with environment-stored credentials. Sensitive keys are never exposed in client-side code.

\subsection{Caching Mechanism}

To enhance efficiency and ensure consistency, \frameworkname\ caches API responses in a PostgreSQL database. This caching mechanism not only reduces the number of repeated API calls, saving time and resources, but also ensures that the ground truth data remains consistent over time. By storing API responses, the platform enables efficient retrieval of previously fetched data, which is particularly valuable when querying the same locations or routes multiple times. The caching mechanism thereby contributes to faster performance and more reliable QA dataset creation, even as real-world map data continues to evolve.

\subsection{Application Scenarios}

MapQaTor is primarily designed to support the creation of both training and evaluation datasets for geospatial question answering (QA), enabling the benchmarking (See Section \ref{sec:mapeval}) and improvement of large language models (LLMs) in geospatial reasoning tasks. In addition to evaluation, MapQaTor can be used to create high-quality training datasets for supervised fine-tuning (SFT) and alignment. Using \frameworknamenospace's extensible architecture, users have the flexibility to evaluate the richness and capabilities of any available map services.

\section{Experiments and Evaluation}



\subsection{Comparison with Manual Methods}


We conducted a controlled experiment to quantify \frameworkname's efficiency gains in geospatial data collection compared to manual methods. Two final-year undergraduate (BSc) students with Google Maps experience performed four geospatial tasks both manually and via \frameworknamenospace. The results (Table \ref{tab:comparison}) demonstrate a significant improvement in data retrieval speed, with \frameworkname requiring at least 30 times less time than the manual approach.

\noindent\textbf{Task Definitions} Four core geospatial operations were evaluated:
\begin{compactitem}
\item \textbf{Place Details}: Retrieve name, address, rating, opening hours, reviews for the Louvre Museum
\item \textbf{Nearby Search}: List 20 nearby restaurants of Louvre Museum, sorted by distance
\item \textbf{Compute Routes}: Generate two alternative driving routes from Eiffel Tower to Louvre Museum
\item \textbf{Search Along Route}: List 20 restaurants along the driving route from Eiffel Tower to Louvre Museum.
\end{compactitem}

\noindent\textbf{Manual Method}
\begin{compactitem}
\item Used Google Maps\footnote{\url{https://www.google.com/maps}} web interface
\item Copied data to spreadsheets with exact formatting
\item Repeated 5 times per task per participant, with the median time recorded to mitigate outliers.
\end{compactitem}

\noindent\textbf{Automated Method}
\begin{compactitem}
\item Executed via \frameworkname's Web Interface
\item Used identical search parameters
\end{compactitem}

\begin{table}[h]
  \centering
  \begin{tabular}{p{3.2cm}|l|l}
    \hline
    \textbf{Task} & \textbf{\frameworkname} & \textbf{Manual} \\
    \hline
    \hline
    Place Details & 10.17 sec & 487 sec \\
    \hline
    Nearby Search & 12.50 sec & 456 sec \\
    \hline
    Compute Routes & 14 sec & 516.5 sec \\
    \hline
    Search Along Route  &  \multirow{1}{*}{15.66 sec} &  \multirow{1}{*}{476 sec}\\ 
    \hline
  \end{tabular}
  \caption{Quantitative comparison between our system and manual methods. }
  \label{tab:comparison}
\end{table}

\subsection{The MapEval Benchmark}\label{sec:mapeval}

To evaluate the annotation quality, we introduce MapEval \cite{dihan2024mapeval}, a benchmark designed to evaluate LLMs on geospatial reasoning tasks. One of its evaluation settings, MapEval-Textual\footnote{\url{https://huggingface.co/datasets/MapEval/MapEval-Textual}}, assesses model performance by prompting LLMs with context and a question, then comparing their responses to the annotated ground truth. This evaluation used 300 MCQs annotated using MapQaTor to benchmark 19 LLMs (e.g., Claude-3.5-Sonnet, GPT-4o, Gemini-1.5-Pro). Preliminary results (Table \ref{tab:mapeval-textual}) reveal significant gaps in model performance on complex spatial tasks, demonstrating the value of MapQaTor in generating high-quality datasets for benchmarking.

MapQaTor's caching mechanism was key in annotating the dataset within the Google Map API’s free tier limit,
while the visualization feature improved annotation accuracy and human evaluation. In MapEval-Textual, two human evaluators, who were not involved in the annotation process, answered the same 300 MCQs, achieving an average accuracy of 86.67\%—more than 20\% higher than the top-performing models (Table \ref{tab:mapeval-textual}). This disparity is attributed to MapQaTor's context visualization feature (Section \ref{sec:visualization}). While LLMs only had access to textual context, lacking visualization capabilities, humans were able to leverage the embedded map to interpret the spatial context.

\begin{table}[h]
  \centering
  \begin{tabular}{l|c}
    \hline
    \textbf{Model} & \textbf{Accuracy (\%)} \\
    \hline
    \hline
    Claude-3.5-Sonnet & 66.33 \\
    \hline
    Gemini-1.5-Pro  & 66.33 \\
    \hline
    GPT-4o  & 63.33  \\
    \hline
    \hline
    Human (with MapQaTor) & 86.67 \\
    \hline
  \end{tabular}
  \caption{MapEval-Textual Performances}
  \label{tab:mapeval-textual}
\end{table}

In MapEval-Textual, LLMs were prompted with \texttt{Formatted Context} (Section \ref{sec:context-optimization}).  Statistics for the 300 MCQs reveal that the average length of Structured Context is 17,534 characters, while the Formatted Context is just 2,536 characters—an 85.54\% reduction. This not only demonstrates MapQaTor's space efficiency but also significantly lowers evaluation costs, as the cost is based on the number of tokens processed.
\section{Related Works}




Recent research has highlighted the potential of map data in mimicking real-world planning tasks through various tools \cite{xie2024travelplanner,zheng2024natural}. Additionally, studies emphasize the significance of caching API call results to establish a stable database for evaluation purposes \cite{guo2024stabletoolbench,xie2024travelplanner}. The development of web-based platforms for integrating geospatial data has also been explored, focusing on streamlining data collection and enhancing the usability of geospatial information for research and development \cite{choimeun2010tool,cai2010summarizing,zheng2014mesa}.

While tool-calling datasets like ToolBench \cite{qin2023toolllm} and APIBank \cite{li2023api} include location-based tasks, their data collection processes lack traceability and reproducibility. This limitation highlights a significant gap in the current landscape: the development of datasets for geospatial question answering is still in its infancy. Existing resources often fail to capture the rich contextual information provided by modern map services. Therefore, there is a pressing need for innovative approaches that effectively leverage the extensive data available from map services to create comprehensive geospatial QA datasets.

 
\section{Conclusion}
In this paper, we have proposed a novel framework, \frameworknamenospace, first of its kind, to automatically fetch rich contextual map service data, which forms the basis to develop language-map benchmark datasets for evaluating SoTA LLMs. Our developed web platform  simplifies data collection for users by offering precise spatial information, user-friendly search, and efficient data retrieval by using Map APIs. Our application also enables user to create geospatial questionnaire. Experimental evaluation suggests that \frameworkname is highly effective in developing geospatial question answer datasets. We believe this approach introduces a new task in geospatial question answering, which has the potential to open a new research direction in the intersection of language models and spatial reasoning.

\section*{Limitations}


Despite the capabilities of \frameworknamenospace, several limitations should be acknowledged. The platform utilizes several paid map APIs, which may incur costs based on usage. During the current public demonstration period, users can explore its features without immediate expenses; however, in the long run, users will need to host the platform independently and integrate their own API keys to access paid functionalities. This requirement necessitates an understanding of the pricing structures associated with the various APIs, potentially impacting accessibility for some users. The platform's functionality is heavily dependent on the availability and stability of external map APIs, meaning that any changes, deprecations, or invalid API keys can negatively impact performance.  The quality of the generated QA pairs is contingent on the retrieved data and users' ability to formulate meaningful questions, which can introduce variability in dataset quality. The evaluation metrics used might not encompass all aspects of usability, possibly overlooking qualitative user feedback. In addition to map service data, other platforms such as Trip Advisor can also be a rich source of additional context for geospatial queries.

\bibliography{acl_latex}
\bibliographystyle{acl_natbib}

\appendix
\newpage
\clearpage

\section{Data Collection Tools}

\begin{figure}[h]
    \centering
    \fbox{\includegraphics[width=1\linewidth]{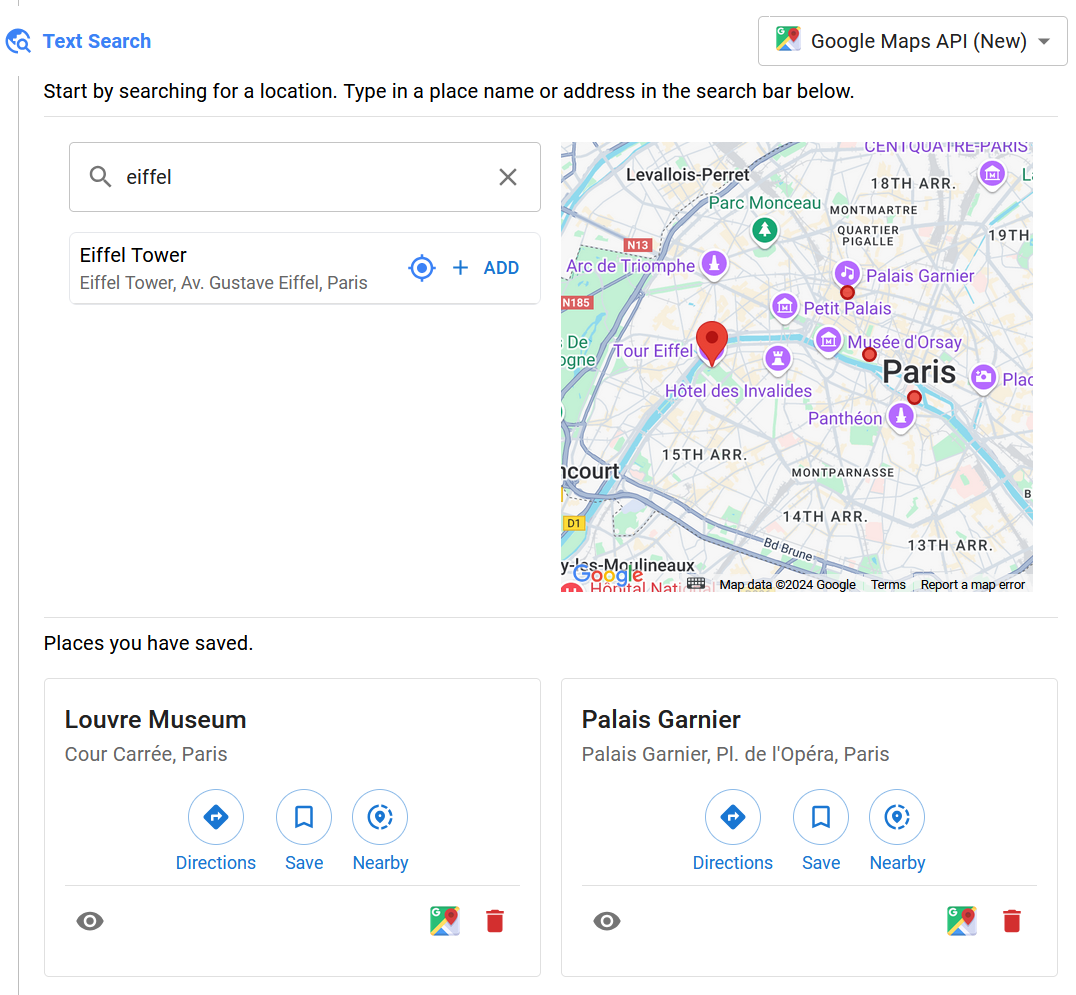}}
    \caption{Search for a place}
    \label{fig:text-search}
\end{figure}

\begin{figure}[h]
    \centering
    \fbox{\includegraphics[width=1\linewidth]{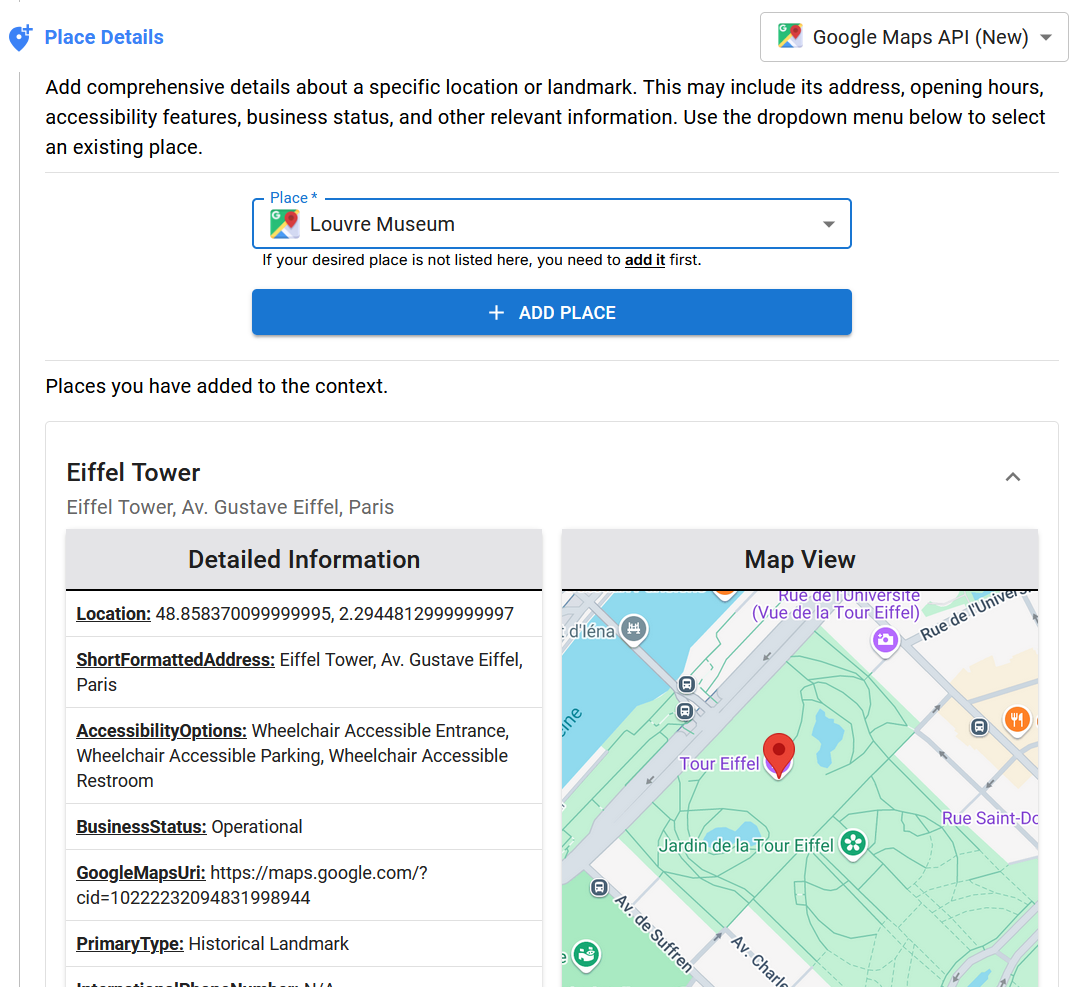}}
    \caption{Fetch full details of a place}
    \label{fig:place-details}
\end{figure}

\begin{figure}[!h]
    \centering
    \fbox{\includegraphics[width=1\linewidth]{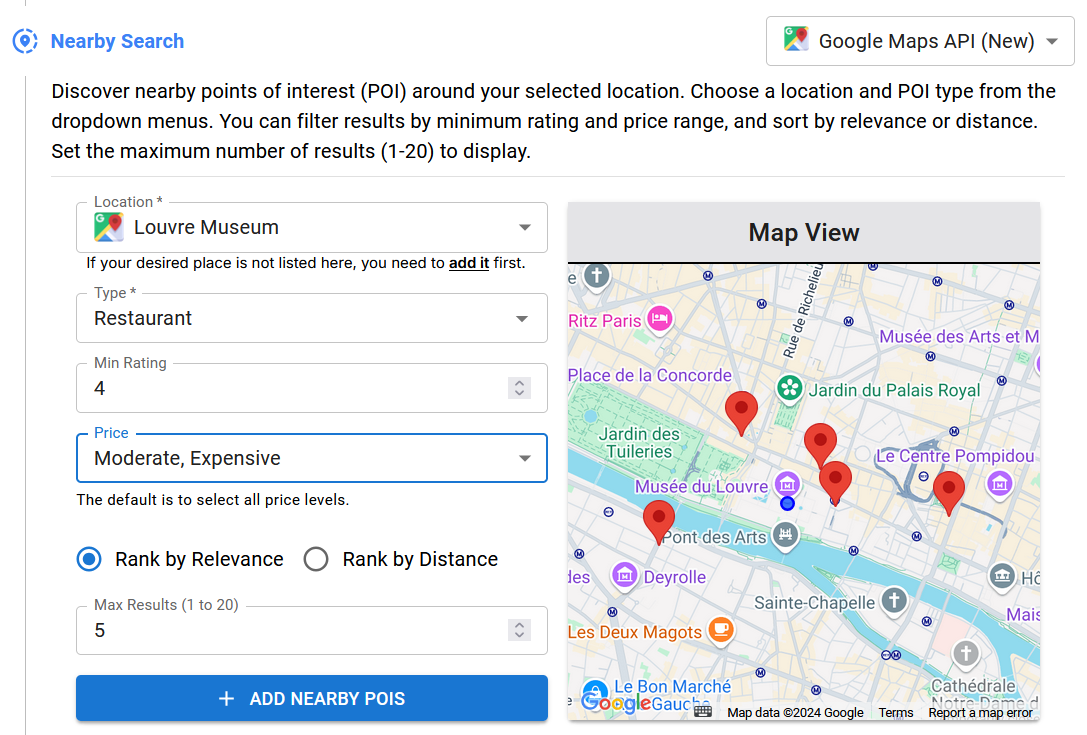}}
    \caption{Search Nearby Places}
    \label{fig:nearby-places-form}
\end{figure}

\begin{figure}[!h]
    \centering
    \fbox{\includegraphics[width=1\linewidth]{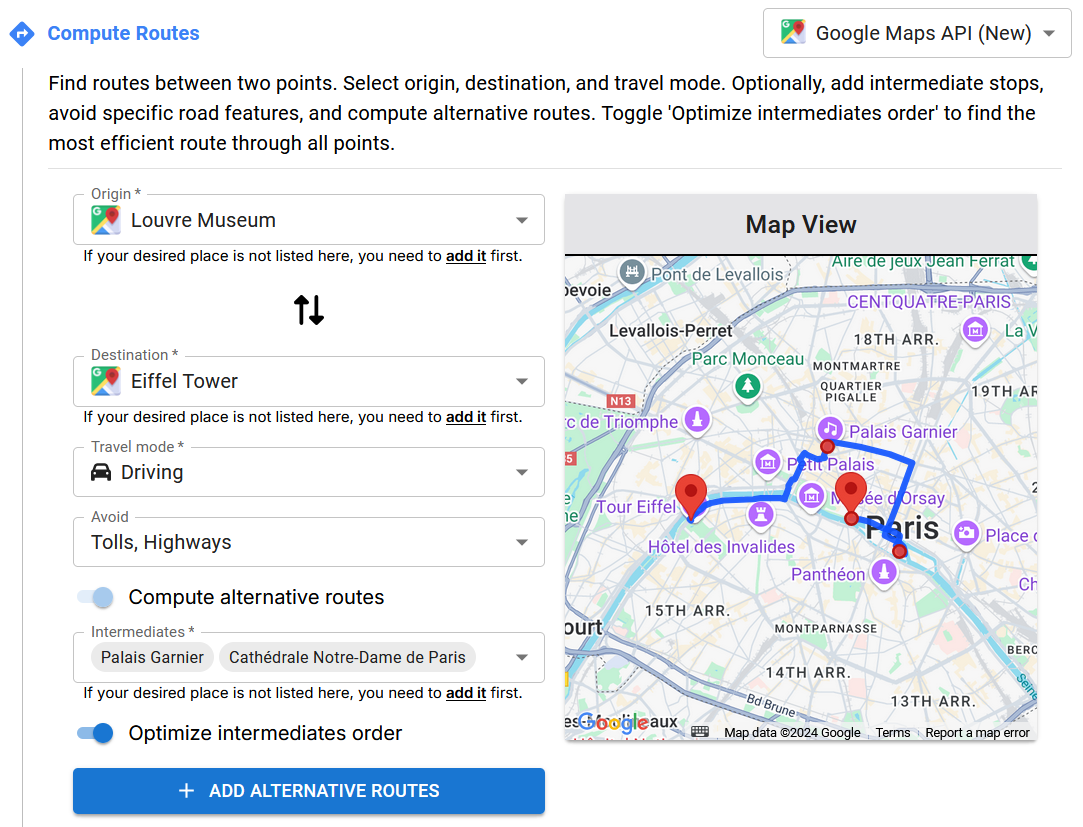}}
    \caption{Find routes between places}
    \label{fig:directions}
\end{figure}

\begin{figure}[!h]
    \centering
    \fbox{\includegraphics[width=1\linewidth]{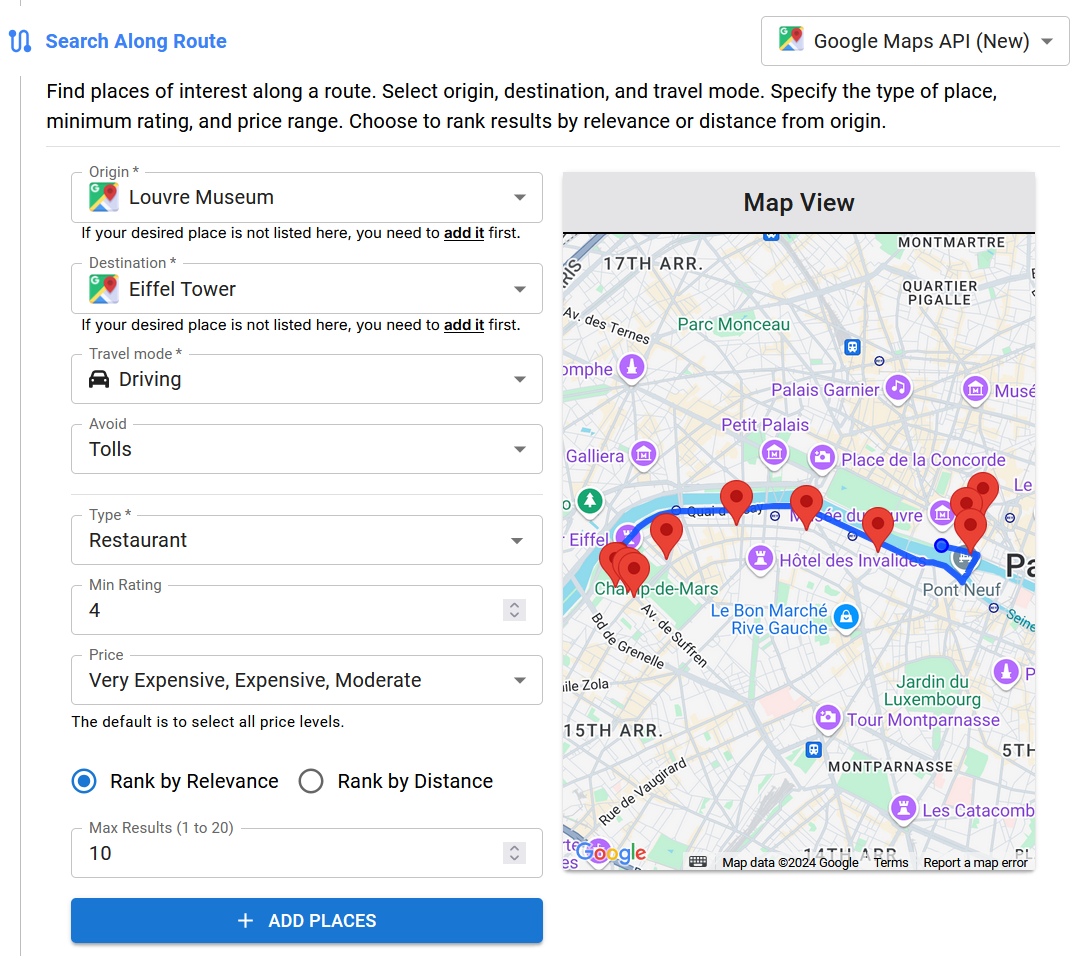}}
    \caption{Search places along a route}
    \label{fig:search-along-route-form}
    \vspace{-5pt}
\end{figure}

\section{Prompt Design Interface} \label{sec:prompt-designer}

\begin{figure*}[h]
\centering
\fbox{\includegraphics[width=1\linewidth]{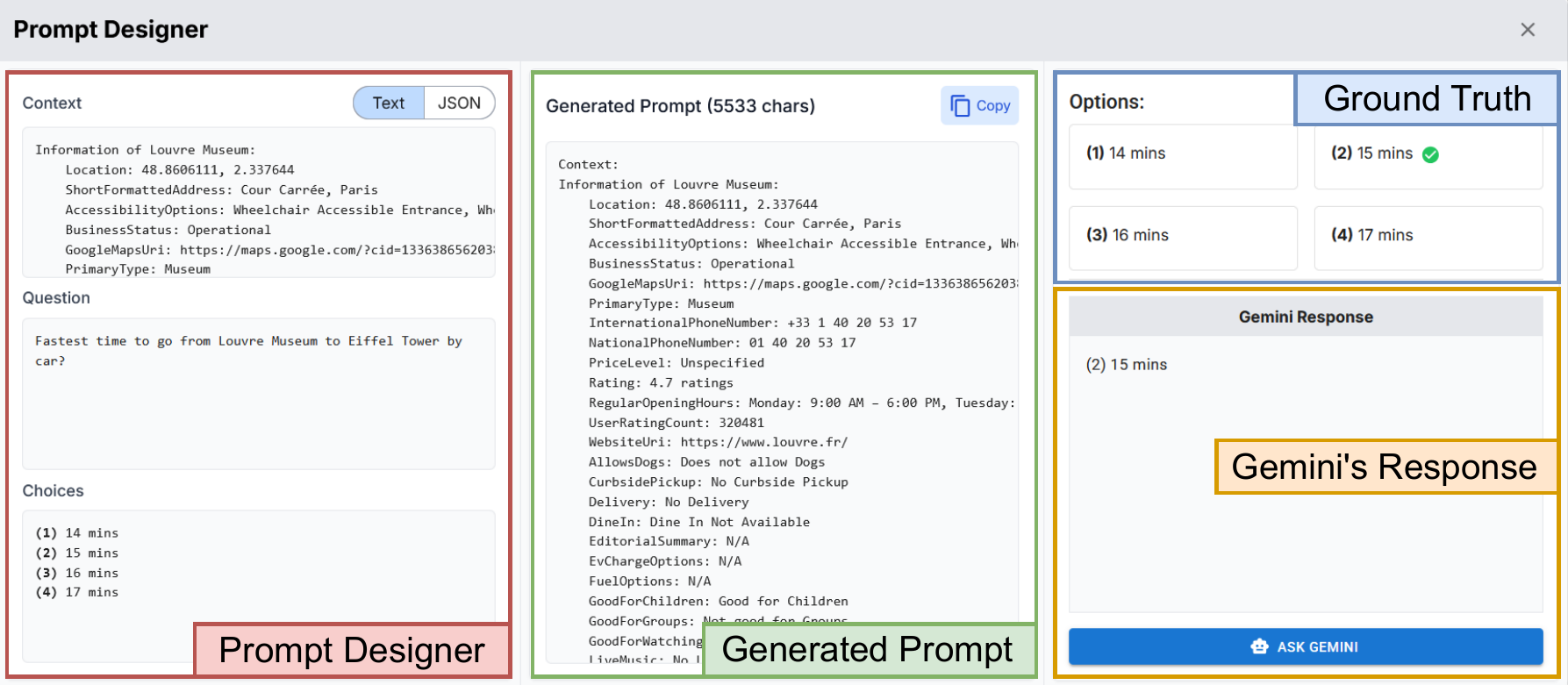}}
\caption{The figure illustrates prompt creation, ground truth comparison, and Gemini's response assessment.}
\label{fig:prompt-designer}
\end{figure*}

The prompt design interface enables users to generate prompts for LLM evaluation by selecting a structured or formatted context. It displays the generated prompt, ground truth answers, and Gemini’s response for comparison. Figure \ref{fig:prompt-designer} illustrates this process.

\section{Exclusion of Temporal Variations in Routing APIs}

To ensure reproducibility, \frameworkname\ removes temporal variations in routing by:

\noindent \textbf{Traffic Awareness Setting:} Routing APIs are set to "TRAFFIC\_UNAWARE," ensuring consistent travel times by ignoring real-time traffic.

\noindent \textbf{Exclusion of Transit Mode:} The "TRANSIT" mode is excluded to prevent variability from schedule changes.

\noindent \textbf{Benefits:}
 \begin{compactitem}
    \item Ensures consistent responses for identical queries.
    \item Focuses evaluations on spatial reasoning, not real-time changes.
    \item Provides a stable baseline for model benchmarking.
\end{compactitem}
These measures enable reliable and reproducible geospatial evaluations in \frameworkname.





\section{API Extension Mechanism}\label{sec:adapter-layer-details}
Figure \ref{fig:tomtom-adapter-layer} demonstrates how new map services are integrated by extending \frameworkname's core tools:



\section{Place Name Suggestion} \label{sec:place-name-suggestion}

Using the TextSearch tool, annotators can retrieve place names. While writing a question or answer, pressing '@' suggests available place names, ensuring consistency between context and QA pairs.

\begin{figure}[h]
\centering
\fbox{\includegraphics[width=1\linewidth]{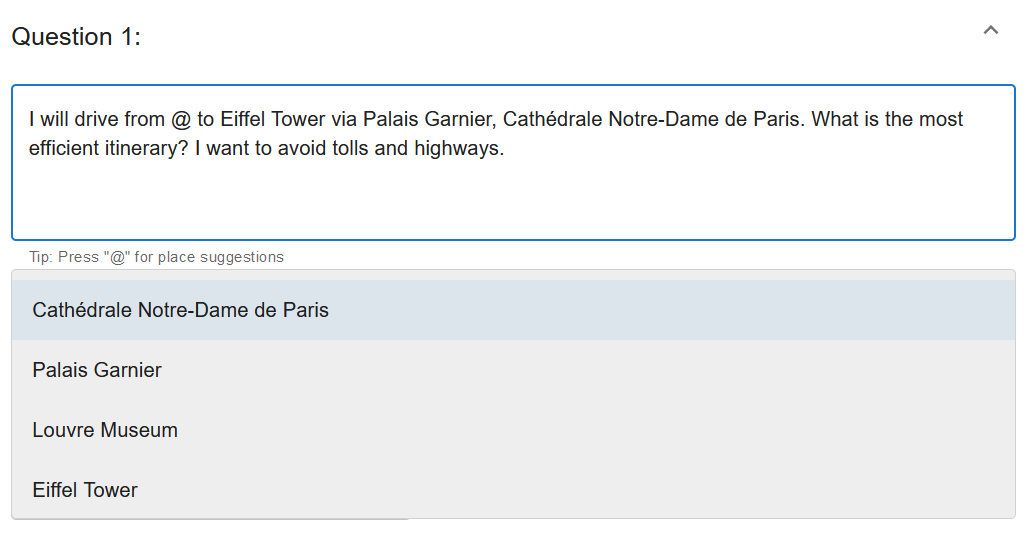}}
\caption{Suggesting available places from the context.}
\label{fig:enter-label}
\end{figure}

\lstset{
    backgroundcolor=\color[RGB]{245,245,245},
    breaklines=true,
    breakindent=0pt,
    basicstyle=\ttfamily\small,
    frame=trbl,
    tabsize=2,
}
\begin{figure}[h]
\begin{lstlisting}
class TomTomApi extends TextSearch {
	constructor() {
		super();
		this.family = "tomtom";
	}

	convertRequest = (query) => {
		return {
			url: "https://api.tomtom.com/search/2/poiSearch/" + query + ".json",
			method: "GET",
			params: {
				key: "key:TOMTOM_API_KEY",
				limit: 5,
				language: "en-US",
			},
		};
	};

	convertResponse = (data) => {
		const places = data.results.map((place) => ({
			id: place.id,
			displayName: {
				text: place.poi.name,
			},
			shortFormattedAddress: place.address.freeformAddress,
			location: {
				latitude: place.position.lat,
				longitude: place.position.lon,
			},
		}));
		return { places };
	};
}
\end{lstlisting}
\caption{Extending Text Search for TomTom API}
\label{fig:tomtom-adapter-layer}
\end{figure}

\end{document}